\documentclass[sigconf]{acmart}

\makeatletter
\def\@acm@checkauthor@country{}
\def\@acm@checkaffil@country{}

\def\@acmDOI{}
\def\@acmISBN{}
\def\@acmConference{}
\def\@acmBooktitle{}
\def\@acmPrice{}

\setcopyright{none}
\acmYear{2025}
\copyrightyear{2025}

\AtBeginDocument{%
  }

\setcopyright{None}
\usepackage[bottom]{footmisc}
\usepackage{verbatim}
\usepackage{todonotes}

\usepackage{appendix} % Package for appendix support
\usepackage{listings} % For code listings
\usepackage{xcolor}
\usepackage{adjustbox}
\usepackage{enumerate}% http://ctan.org/pkg/enumerate
\usepackage{hyperref}

\usepackage{float}
\restylefloat{table}
\usepackage{placeins}
\usepackage{stfloats}
\usepackage{afterpage}

\usepackage{fancyvrb}
\usepackage{fvextra}

\usepackage{xcolor}         % colors

\begin{document}

%%
%% The "title" command has an optional parameter,
%% allowing the author to define a "short title" to be used in page headers.
%\title{Hybrid Retrieval Augmented Generation: LiveRAG Challenge}
\title[Hybrid RAG with Dynamic Test Sets]{Evaluating Hybrid Retrieval Augmented Generation using Dynamic Test Sets: LiveRAG Challenge}

%%
%% The "author" command and its associated commands are used to define
%% the authors and their affiliations.
%% Of note is the shared affiliation of the first two authors, and the
%% "authornote" and "authornotemark" commands
%% used to denote shared contribution to the research.

\author{Chase Fensore$\dagger$, Kaustubh Dhole$\dagger$, Joyce Ho$^{*}$, Eugene Agichtein$^{*}$}
\affiliation{%
%\institution{Department of Computer Science}
  \institution{Emory University, USA}
  % \country{USA}
  }
\email{{cfensore, kdhole}@emory.edu}

%%
%% By default, the full list of authors will be used in the page
%% headers. Often, this list is too long, and will overlap
%% other information printed in the page headers. This command allows
%% the author to define a more concise list
%% of authors' names for this purpose.
%\renewcommand{\shortauthors}{Fensore et al.}

% Commenting colors:
\newcommand{\chase}[1]{{\color{cyan}[Chase: #1]}} 
\newcommand{\kaustubh}[1]{{\color{darkpink}[Kaustubh: #1]}} 

\newcommand{\joyce}[1]{{\color{orange}[Joyce: #1]}} 
\newcommand{\eugene}[1]{{\color{darkgreen}[Eugene: #1]}}

%%
%% The abstract is a short summary of the work to be presented in the
%% article.

\begin{abstract}
%In this article, we present our submission to the LiveRAG Challenge 2025 along with offline experiments, which employ a hybrid retrieval-augmented generation (RAG) approach. We detail our experimental evaluation across various components, including query formulation, retrieval, ranking, and generation. 
% Additionally, we discuss offline experiments demonstrating that incorporating a re-ranker improved performance by X compared to our submitted system.  
We present our submission to the LiveRAG Challenge 2025, which evaluates retrieval-augmented generation (RAG) systems on dynamic test sets using the FineWeb-10BT corpus. Our final hybrid approach combines sparse (BM25) and dense (E5) retrieval methods and then aims to generate relevant and faithful answers with Falcon3-10B-Instruct. Through systematic evaluation on 200 synthetic questions generated with DataMorgana across 64 unique question-user combinations, we demonstrate that neural re-ranking with RankLLaMA improves MAP from 0.523 to 0.797 (52\% relative improvement) but introduces prohibitive computational costs (84s vs 1.74s per question). While DSPy-optimized prompting strategies achieved higher semantic similarity (0.771 vs 0.668), their 0\% refusal rates raised concerns about over-confidence and generalizability. Our submitted hybrid system without re-ranking achieved 4\textsuperscript{th} place in faithfulness and 11\textsuperscript{th} place in correctness among 25 teams. Analysis across question categories reveals that vocabulary alignment between questions and documents was the strongest predictor of performance on our development set, with document-similar phrasing improving cosine similarity from 0.562 to 0.762.

\end{abstract}

\maketitle
%\received[revised]{12 March 2009}
%\received[accepted]{29 May 2025}

%%
%% This command processes the author and affiliation and title
%% information and builds the first part of the formatted document.

%%%%%%%%%%%%%%%%%%%%%%%%%%%%%%

\section{Introduction}
%Retrieval-Augmented Generation (RAG) has become increasingly important for improving the factual accuracy and reliability of responses generated by Large Language Models (LLMs)~\cite{lewis2020retrieval, izacard2022atlas}. However, the rapidly growing reasoning capabilities of LLMs have begun to render existing static benchmarks less effective in accurately assessing RAG performance in practical scenarios, emphasizing the need for more dynamic Question-Answering (QA) benchmarks. To address this challenge, LiveRAG\footnote{\url{https://liverag.tii.ae/}} organized a competition inviting submissions aimed at developing adaptive QA systems. In this paper, we describe our submission to LiveRAG, outlining our system ______, detailing baseline experiments, the motivations behind our key design decisions, and presenting preliminary evaluation results.
%%%%%%%%%%%%%%%%%%%%%%%%%%%%%%
%\section{Background}
The LiveRAG Challenge 2025 presents a unique evaluation framework for retrieval augmented generation (RAG) systems using dynamic test sets that capture the evolving nature of real-world information needs. Unlike traditional static benchmarks, this challenge evaluates systems on their ability to handle diverse question types, user expertise levels, and rapidly changing information landscapes using the FineWeb-10BT corpus~\cite{penedo2024fineweb}.

Our approach focuses on developing a hybrid retrieval system that combines the complementary strengths of sparse and dense retrieval methods. Sparse retrieval excels at exact term matching and handling specific factual queries, while dense retrieval captures semantic similarity and contextual understanding. By combining these approaches and exploring document reformulation, neural re-ranking, and prompt optimization, we ultimately aimed to maximize correctness and faithfulness of our RAG system's responses on the unseen LiveRAG Challenge Day test set. 
% maximize retrieval coverage and precision.
% No: aim to maximize Correctness & Faithfulness (challenge metrics).

The key contributions of our work include: (1) a systematic evaluation of hybrid retrieval strategies on synthetic dataset we generated with the DataMorgana tool \citep{datamorgana_filiceGeneratingDiverseQA2025} that mirrors real-world QA diversity, (2) analysis of performance across different question and user categories as defined by DataMorgana, (3) demonstration of the effectiveness of neural re-ranking in improving RAG system performance, which also comes at high computational cost, and (4) insights into trade-offs between development set optimization and real-world robustness through DSPy~\cite{khattabDSPyCompilingDeclarative2023} prompt optimization experiments.

This paper is organized as follows: In Section~\ref{sec:methds}, we first describe our retrieval and RAG pipeline in detail. In Section~\ref{sec:exp}, we demonstrate our experiments and discuss our results.

%%%%%%%%%%%%%%%%%%%%%%%%%%%%%%
\section{Methods}
\label{sec:methds}
We now describe our baseline experiments carried out to inform the development of our final RAG system.

\subsection{Data Generation}
% Document-QA Pair Generation
We utilized DataMorgana to create a development QA dataset (n=200) that reflects the diversity expected in real-world RAG applications. Our synthetic dataset incorporated multiple question categories and user categorizations, to simulate up to 64 unique combinations of RAG QA settings.
\newline
\textbf{Question Categories}: We used four question categorizations consistent with those introduced by DataMorgana: \citep{datamorgana_filiceGeneratingDiverseQA2025}
\begin{enumerate}
    \item \textbf{Question Factuality}: Balancing factoid questions (50\%) seeking specific information with open-ended questions (50\%) encouraging detailed responses
    \item \textbf{Question Premise}: Including direct questions (50\%) without user context and premise-based questions (50\%) where users provide relevant background
    \item \textbf{Question Phrasing}: Covering concise natural questions (25\%), verbose natural questions (25\%), short search queries (25\%), and long search queries (25\%)
    \item \textbf{Question Linguistic Variation}: Distinguishing between questions using document-similar terminology (50\%) and document-distant phrasing (50\%)
\end{enumerate}

\textbf{User Categories}: We focused on \textbf{User Expertise} categories, with expert users (80\%) asking complex questions and novice users (20\%) asking basic questions.

\subsection{Retrieval Architecture}
Documents were first split into 512 token chunks with a sentence-aware splitter. For the initial stage of retrieval, we explored three retrieval strategies:
\begin{enumerate}
    \item \textbf{Sparse Retrieval}: OpenSearch BM25 based index. k=30 documents
    \item \textbf{Dense Retrieval}: Pinecone with E5 embeddings (\verb|intfloat/| \newline \verb|e5-base-v2|, size 768) and cosine similarity for retrieval. k=30 documents
    \item \textbf{Hybrid Retrieval}: Combined sparse and dense retrieval (k=30 each), selecting top 10 documents based on normalized score fusion~\citep{lee_combining_1995}
\end{enumerate}
%\chase{Discuss how we chose k=30 for each, and final k=10. What train set did we use to determine these?}

% \todo[inline]{How was the overall score computed? Was any exploration done on merging algorithm/options - eg., weighted sum vs. interleaving vs. reciprocal rank vs. product, etc - any observations on which type of retrieval was more useful/effective for this task?}

\textbf{Document Reformulation}
\newline
We explored doc2query-enhanced sparse indexing~\cite{nogueira2019document}, where documents were augmented with generated questions to improve retrieval coverage. We created a custom doc2query-enhanced sparse BM25 index, using 512 token chunks similar to the OpenSearch BM25 index. When using this index, the top 30 documents were retrieved and included in the final prompt. However, this doc2query approach showed limited effectiveness in our experiments. 

% \todo[inline]{Need a bit more info. Which embeddings used for dense index, it seems like 2 separate indexes with late merging? please say a bit more about implementation/libraries used for each, doesn't have to be alot.}

\textbf{Generative Re-ranking}
\newline
We employed RankLLaMA-7B (\verb|castorini/rankllama| \newline  \verb|-v1-7b-lora-passage|)~\cite{rankllama} for pointwise re-ranking of the top k retrieved passages --- this means that the a given retrieved passage is assigned a score by the generative re-ranker independently of all other retrieved passages. Pointwise re-rankers have shown improved retrieval effectiveness in many tasks and are more robust compared to other approaches like listwise.

To do this, the re-ranker uses the prompt structure from the \verb|pyterrier_genrank| package~\cite{Dhole_PyTerrier_Genrank}, where each prompt contains exactly one query and one passage (Figure \ref{fig:rerank_prompt_total}).

\begin{figure}[htbp]
\begin{minipage}[b]{0.5\textwidth} % Full width for the first minipage
\begin{tiny}
\begin{Verbatim}[frame=single, commandchars=\\\{\},breaklines=true, breakanywhere=true, breaksymbol=, breakanywheresymbolpre=]
Query: \{query\} Passage: \{passage\}
\end{Verbatim}
\end{tiny}
\end{minipage}
\caption{Pointwise Re-ranking Prompt}
\label{fig:rerank_prompt_total}
\end{figure}

% \todo[inline]{Which prompt was used for re-ranking? Were variants compared/explored?}

\subsection{Answer Generation}
% With Falcon-10B-Instruct

We used the \verb| Falcon3-10B-Instruct |\cite{Falcon3} model for answer generation with temperature=0.6 and top\_p=0.9. We performed manual prompt engineering and arrived at a base prompt aimed to emphasize conciseness and faithfulness (Figure \ref{fig:prompt_total}).

\begin{figure}[t]
\begin{minipage}[b]{0.5\textwidth} % Full width for the first minipage
\begin{tiny}
\begin{Verbatim}[frame=single, commandchars=\\\{\},breaklines=true, breakanywhere=true, breaksymbol=, breakanywheresymbolpre=]
"""You are an AI assistant tasked with answering questions based on the provided information.
        
Information: 
{context as k chunks/documents}

Question: {query}

Answer the question based only on the provided information. Keep the answer concise, limited to 200 tokens. If the information doesn't contain the answer, say "I don't have enough information to answer this question."

Answer:"""
\end{Verbatim}
\end{tiny}
\end{minipage}
\caption{Answer Generation Prompt}
\label{fig:prompt_total}
\end{figure}

Few-shot and Chain-of-Thought (CoT) approaches were also explored via \verb| DSPy | \cite{khattabDSPyCompilingDeclarative2023}. For few-shot prompting using DSPy, we used a 160/40 train/validation split of our synthetic dataset. Answers were scored during prompt optimization with question similarity measured by MiniLM.\footnote{\url{https://huggingface.co/sentence-transformers/all-MiniLM-L6-v2}\label{minilm}} MIPROv2 and BootstrapFewShot optimizers were studied for CoT and few-shot respectively. For CoT prompting, we explored MIPROv2 to optimize the reasoning instructions and step-by-step decomposition structure, producing prompts that aim to explicitly guide the model through Context → Question → Reasoning → Answer sequences. BootstrapFewShot automatically curates high-quality demonstrations by bootstrapping examples from our training set, selecting demonstrations that maximize performance on our evaluation metric. This technique generates few-shot prompts containing up to four optimally chosen examples that are most representative of successful question-answering patterns in our domain. The optimization process operates on a subset of our 200-question synthetic dataset, using an 80/20 train-dev split with cross-validation. The resulting optimized prompt templates are serialized as JSON configurations and integrated into our generation pipeline, allowing dynamic selection between default prompting, DSPy-optimized CoT reasoning, and DSPy-optimized few-shot demonstration strategies.

% The real formula i used for enhanced_rag_metric (DSPy train scoring): 
% final_score = 0.2 * exact_match_score + 0.4 * relevance_score + 0.4 * faithfulness_score

% \todo[inline]{This part is interesting, please add more details: prompts, how examples were chosen, any other info to give the reader idea on how the method works and was there some optimization/exploring done?}

% \chase{We explored various prompting techniques, including few-shot and chain-of-thought reasoning using DSPy, but found that the simple prompt performed best within our time constraints while maintaining low refusal rates. }

\section{Experimental Results}\label{sec:exp}
\subsection{Retrieval Performance Analysis}
Table \ref{tab:retr_metrics} presents our retrieval, document reformulation, and re-ranking results on our DataMorgana synthetic dataset. The hybrid approach matches sparse retrieval performance in most metrics, suggesting that the dense component provides complementary rather than additive benefits. However, the combination creates a more robust foundation for re-ranking and shows stronger downstream answer generation performance (Table 
\ref{tab:gen_metrics}).
The most significant finding is the substantial improvement achieved by neural re-ranking. RankLLaMA re-ranking improved MAP from 0.523 using Hybrid to 0.797, representing a 52\% relative improvement. 
% This suggests that while initial retrieval provides reasonable recall, the ranking quality significantly impacts downstream generation performance.
Interestingly, doc2query enhancement performed poorly across all retrieval metrics, possibly due to the domain mismatch between the enhancement model's training data and the FineWeb-10BT corpus characteristics.
Despite the impressive retrieval and generation performance boost from including RankLLaMA, this approach was computationally expensive (\textasciitilde{}84 seconds/question). Applying this solution to the 500-question test set on LiveRAG Challenge Day would have been infeasible as there was a 2 hour window to generate answers, and this generative re-ranking strategy would have required \textasciitilde{}12 hours. As a result, we opted to use the hybrid retrieval approach for the LiveRAG Challenge Day, as it gave the second strongest retrieval and downstream generation performance on our development question set while maintaining low computational overhead (\textasciitilde{}1.74 seconds/question).

\begin{table*}[ht]
\centering
\caption{\smaller{Baseline retrieval, re-ranking results on DataMorgana synthetic QA dataset (n=200). \textbf{Bold} denotes best, \underline{underline} denotes second best. Time indicates the mean time required for retrieval and generation, in seconds.}} % NOTE: generation took 1.55 s on average.
\label{tab:retr_metrics}
\resizebox{\textwidth}{!}{%
\begin{tabular}{l|ccccccc|c}
\toprule
\textbf{Name} & \textbf{MAP} & \textbf{Recip. Rank} & \textbf{nDCG@10} & \textbf{Recall@1} & \textbf{Recall@10} & \textbf{Prec@1} & \textbf{Prec@10}  & \textbf{Time (s)} \\
\midrule
Sparse (OpenSearch BM25) & \underline{.523} & \underline{.347} & \underline{.497} &  \underline{.285} & \underline{.485} & \underline{.285} & \underline{.074} & 1.57 \\ % python main.py --input 200QAs_DMinput.jsonl --output sparse_only_k30_final10_200DM.json --use-aws --retriever-type sparse --top-k-sparse 30 --top-k-final 10 --falcon-mode ai71 --ai71-api-key "..."
Dense (Pinecone E5) & .352 & .260 & .367 & .190 & .435 & .190 & .058  & 1.56 \\ % python main.py --input 200QAs_DMinput.jsonl --output dense_only_k30_final10_200DM.json --use-aws --retriever-type dense --top-k-dense 30 --top-k-final 10 --falcon-mode ai71 --ai71-api-key "..."
Hybrid & \underline{.523} & \underline{.347} & \underline{.497} & \underline{.285} & \underline{.485} & \underline{.285} & \underline{.074}  & 1.74 \\

\hline
Hybrid $\rightarrow$ RankLLaMA & \textbf{.797} & \textbf{.422} & \textbf{.710} & \textbf{.340} & \textbf{.590} & \textbf{.340} & \textbf{.116}  & 84.37 \\

\hline
doc2query+BM25 & .321 & .321 & .353 & .275 & .455 & .275 & .046  & 2.93 \\

\bottomrule
\end{tabular}%
}
\end{table*}

\subsection{Generation Performance Analysis}
Table \ref{tab:gen_metrics} shows the automatic evaluation of generated answers across different retrieval strategies. We used multiple metrics to capture different aspects of answer quality: ROUGE scores for measuring n-gram recall~\cite{lin2004rouge}, BLEU for precision-oriented matching~\cite{papineni2002bleu}, cosine similarity of MiniLM embeddings\footref{minilm} to measure semantic match. We also measure the refusal rate to measure system reliability, by comparing whether the answer contains any of a set of refusal responses like ``not enough information.''

\begin{table}[ht]
\centering
\caption{\smaller{Generation evaluation on DataMorgana synthetic QA dataset (n=200). Semantic similarity was calculated with cosine similarity of embeddings from MiniLM-L6-v2. \textbf{Bold} denotes best, \underline{underline} denotes second best.}}
\label{tab:gen_metrics}
\resizebox{\columnwidth}{!}{%
\begin{tabular}{l|ccccc}
\toprule
\textbf{Name} & \textbf{ROUGE-1} & \textbf{ROUGE-L} & \textbf{BLEU} & \textbf{Cos. Sim.} & \textbf{\% Refusal ($\downarrow$)} \\
\hline
\textbf{Retrieval} & & & & & \\
Sparse & .366 & .276 & .115 & .659 & 17.50\% \\

Dense & .337 & .244 & .079 & .668 & 15.00\% \\

Hybrid & \underline{.368} & \underline{.279} & \underline{.117} & .668 & 17.00\% \\

\hline
\textbf{Re-ranking} & & & & & \\
Hybrid $\rightarrow$ RankLLaMA & \textbf{.403} & \textbf{.307} & \textbf{.123} & .754 & \underline{3.50}\% \\

\hline
\textbf{Document Reformulation} & & & & & \\
doc2query+BM25 & .337 & .247 & .088 & .641 & 19.50\% \\

\hline
\textbf{Prompting} & & & & & \\ 
Few-shot (w/ Hybrid)& \underline{.368} & .266 & .106 & \textbf{.771} & \textbf{0.00}\% \\ % TIME: 2.35s, avg length: 118.1 words
CoT (w/ Hybrid)& .358 & .255 & .096 & \underline{.756} & \textbf{0.00}\% \\ % TIME: 2.54s, avg length: 117.8 words.

%\textbf{Passage Re-structuring} & & & & & \\ % Content-Aware Passage Structuring: Implement for hybrid: After ranking, structure retrieved passages using HTML/markdown, providing Falcon with better cues about document structure and importance hierarchies.
\bottomrule
\end{tabular}%
}
\end{table}
% TODO - compare mean answer length -> ground truth mean answer length.

Our re-ranker-based hybrid approach consistently outperforms all other methods across all metrics. Most notably, the refusal rate drops from 17\% to 3.5\%, indicating that better retrieval quality leads to more confident and more accurate answer generation, at least according to automatic evaluation. The cosine similarity improvement from 0.668 to 0.754 suggests that re-ranking helps surface semantically relevant content that better supports accurate answer generation.
Hybrid retrieval also showed strong generation performance with the second highest ROUGE-1, ROUGE-L, and BLEU.

\textbf{Advanced Prompting Analysis}
\newline
The DSPy-optimized prompting strategies show interesting but concerning patterns. Few-shot prompting achieved the highest semantic similarity (0.771 cosine similarity) and CoT prompting showed strong performance (0.756), both significantly outperforming our hybrid approach with a manually crafted prompt (0.668). However, both advanced prompting methods exhibit 0\% refusal rates, indicating potential over-confidence that could be problematic for generalization.

While the improved semantic similarity suggests these methods generate more semantically coherent responses on our development set, the complete absence of refusal responses raised concerns about calibration and robustness. In RAG systems, appropriate refusal when information is insufficient is crucial for maintaining trustworthiness, especially given the LiveRAG Challenge criteria scoring for incorrect answers vs refusals. The 0\% refusal rate suggests these optimized prompts may be generating confident but potentially incorrect responses when faced with insufficient context.

This over-optimization phenomenon aligns with known challenges in prompt optimization, where methods can achieve high performance on development metrics while degrading real-world robustness~\cite{opsahl-ong_optimizing_2024, soylu_fine-tuning_2024}. Given these concerns about generalizability, we opted to use our conservative baseline prompting strategy for the LiveRAG Challenge submission (Figure \ref{fig:prompt_total}), which maintains appropriate refusal behavior while achieving strong faithfulness rankings.

\subsection{Performance by Question, User Categories}
Table \ref{tab:subgroup_perf} analyzes generation performance stratified by DataMorgana's question and user categories using our best-performing hybrid approach \textit{without} re-ranking. Based on Table \ref{tab:subgroup_perf}, we observe significant performance variations across question and user categories using our hybrid retrieval approach.

\textbf{Question Characteristics}: Factoid questions substantially outperform open-ended questions (ROUGE-1: 0.407 vs 0.332, BLEU: 0.162 vs 0.075), confirming that the system excels at extracting specific information but struggles with comprehensive explanatory responses. Direct questions slightly outperform premise-based questions in semantic quality (0.683 vs 0.652 cosine similarity) with lower refusal rates (14.7\% vs 19.4\%).

\textbf{Phrasing Impact}: Natural language formulations significantly outperform search query formats. Verbose natural questions achieve the highest semantic similarity (0.739) and lowest refusal rate (8.9\%), while short search queries perform poorly across metrics (0.567 cosine similarity, 26.1\% refusal rate). This demonstrates clear sensitivity to query formulation quality.

\textbf{Vocabulary Alignment}: The most pronounced performance gap occurs between questions using document-similar versus distant terminology. Similar phrasing substantially improves performance (ROUGE-1: 0.431 vs 0.296, cosine similarity: 0.762 vs 0.562), with refusal rates dropping from 25.5\% to 9.4\%. This highlights vocabulary matching as a critical factor in RAG effectiveness.

\textbf{User Expertise}: Our system consistently outperforms on expert-level questions relative to novice questions (cosine similarity: 0.709 vs 0.628), likely reflecting the more precise, focused formulations and vocabulary present in expert-level questions.

\textbf{Extreme Case Analysis}: 
Looking at combinations of question and user characteristics, the primary differentiators for performance of our hybrid RAG system were Question Premise ($\uparrow$ direct, $\downarrow$ with-premise) and Question Phrasing ($\uparrow$ concise+natural, $\downarrow$ short search query). These sets of the best and worst performing question clusters had identical settings for all other characteristics: Question Factuality (open-ended), Linguistic Variation (similar to document), and User Expertise (novice). Our best combination achieves 0.822 cosine similarity with zero refusal rate, while the worst combination shows 0.479 cosine similarity with 33.3\% refusal rate.

\begin{table}[ht]
\centering
\caption{\smaller{Breakdown of answer quality stratified by DataMorgana QA characteristics. Shows hybrid retrieval + basic prompting approach, for practicality. Bold indicates best performance, excluding best and worst combination rows.}}
\label{tab:subgroup_perf}
\resizebox{\columnwidth}{!}{%
\begin{tabular}{lccccc}
\toprule
\textbf{Name (n)} & \textbf{ROUGE-1} & \textbf{ROUGE-L} & \textbf{BLEU} & \textbf{Cos. Sim.} & \textbf{\% Refusal ($\downarrow$)}  \\
\hline
\textbf{Question Category} & & & & & \\
\hline
\textbf{Question Factuality} & & & & & \\
\hspace{3mm}Factoid (96) & .407 & .334 & \textbf{.162} & .687 & 16.67\% \\
\hspace{3mm} Open-ended (104) & .332 & .229 & .075 & .651 & 17.30\% \\
\textbf{Question Premise} & & & & & \\
\hspace{3mm} Direct (102) & .372 & .287 & .123 & .683 & 14.70\% \\
\hspace{3mm} With-premise (98) & .363 & .272 & .110 & .652 & 19.40\% \\

\textbf{Question Phrasing} & & & & & \\
\hspace{3mm} Concise, Natural (54) & .370 & .304 & .127 & .674 & 18.50\% \\
\hspace{3mm} Verbose, Natural (56) & .414 & .298 & .135 & .739 & \textbf{8.90\%} \\
\hspace{3mm} Short Search Query (46) & .312 & .233 & .084 & .567 & 26.10\% \\
\hspace{3mm} Long Search Query (44) & .365 & .274 & .114 & .677 & 15.90\% \\

\textbf{Linguistic Variation} & & & & & \\
\hspace{3mm} Similar to Document (106) & \textbf{.431} & \textbf{.336} & .153 & \textbf{.762} & 9.40\% \\
\hspace{3mm} Distant from Document (94) & .296 & .216 & .075 & .562 & 25.50\% \\

\hline
\textbf{User Category} & & & & & \\ 
\hline
\textbf{User Expertise} & & & & & \\
\hspace{3mm} Novice (101) & .336 & .252 & .099 & .628 & 20.8\% \\
\hspace{3mm} Expert (99) & .401 & .308 & .135 & .709 & 13.1\% \\

\bottomrule
Best Combination (5) & .328 & .254 & .083 & .822 &  0.0\% \\
Worst Combination (6) & .259 & .174 & .037 & .479 & 33.3\% \\
%\textbf{Passage Re-structuring} & & & & & \\ % Content-Aware Passage Structuring: Implement for hybrid: After ranking, structure retrieved passages using HTML/markdown, providing Falcon with better cues about document structure and importance hierarchies.
\bottomrule
\end{tabular}%
}
\end{table}

%Primary: Average Cosine Similarity (semantic quality)
%Secondary: Combination must have ≥5 samples for reliability
%Best: combinations with highest cosine similarity

\subsection{LiveRAG Challenge Performance}
In the preliminary LiveRAG Challenge evaluation on 500 unseen test questions, our hybrid RAG system achieved rankings of 11\textsuperscript{th} place for correctness (11/25) and 4\textsuperscript{th} place for faithfulness (4/25) among participating teams. The strong faithfulness ranking suggests that our emphasis on retrieval quality and conservative answer generation effectively grounded responses in the provided context, while the moderate correctness ranking indicates room for improvement in answer accuracy. The performance gap between faithfulness (4\textsuperscript{th}) and correctness (11\textsuperscript{th}) suggests our conservative prompting strategy successfully grounded responses in retrieved context but may have been overly cautious, leading to refusal responses for several questions. This aligns with our synthetic evaluation showing 17\% refusal rates for the hybrid approach. 

\subsection{Limitations}
Our approach has several limitations. First, the hybrid retrieval method showed minimal improvement over sparse retrieval alone, suggesting dense retrieval may not provide complementary benefits for this corpus. Second, our conservative answer generation strategy, while improving faithfulness, may have reduced correctness by refusing to answer questions with sufficient but not obvious evidence. Finally, because our experiments focused on time-consuming stages like prompt optimization with DSPy and generative re-ranking, we chose to use a relatively small development set (n=200) compared to the 10,000 DataMorgana credits allotted to teams. This small development set size may have limited our systems generalizability to the test set for the LiveRAG Challenge,

\section{Conclusion}
We presented a hybrid RAG approach combining sparse and dense retrieval for the LiveRAG Challenge 2025. Our systematic evaluation using DataMorgana's 64 question-user combinations reveals that vocabulary alignment is the most critical factor for RAG performance, with neural re-ranking providing substantial improvements at prohibitive computational cost. While our submitted system achieved strong faithfulness rankings (4\textsuperscript{th}/25), the moderate correctness performance (11\textsuperscript{th}/25) highlights the tension between conservative and accurate answer generation.

Despite the strong performance of the generative re-ranker in our development process, we determined it was computationally infeasible for the competition timing constraints. Building computationally efficient generative re-rankers is an active area of study. Works like RankZephyr~\cite{pradeep_rankzephyr_2023} and RankGPT~\cite{sun_is_2024} have studied listwise generative re-ranking for challenging queries requiring context from multiple passages, however listwise re-ranking is often slower than pointwise re-rankers. In addition to using batching to increasing feasibility of generative re-rankers in practice, we encourage future work to focus on strategies to increase efficiency of generative re-rankers like RankLLaMA, as they show strong performance in this general-purpose RAG QA task but are computationally costly.

Future work could also focus on adaptive retrieval strategies that adjust computational overhead based on question complexity, and more sophisticated answer generation that balances faithfulness with completeness. The substantial performance variations across question categories (0.479 to 0.822 cosine similarity) suggest that question-aware system design could yield significant improvements.

%%%%%%%%%%%%%%%%%%%%%%%%%%%%%%%%%%%%%%%%%%%%
% REQUIRED TOPICS TO INCLUDE (from LiveRAG admin): 
% Some guidelines that might help you in writing your final report.
%Please do not use an LLM for that -- we would like to hear about your system in your own words.
% Your report should cover the main work and experiments done for the Challenge. This may include data generation, query rewriting, RAG implemented architecture, prompt generation, answer post-processing, and others. Please elaborate mostly on the innovative parts of your work and on your interesting findings.
% Expand on the usage of DataMorgana (the question/user categories you tried, etc.), or in case you used other synthetic data generation tools. Explain how you measured performance and how you improved your system.
% Describe your retrieval system. Was it based on the pre-built indices or did you build your own search index? Did you try Sparse/Dense/Hybrid/Other approaches? Did you re-rank the search results?
% Describe the prompt used for answer generation. How was it built and why?
%Did you modify the Falcon parameters? If so elaborate which ones and how did you set their value? Elaborate on your experience working with AI71 Falcon service.
% Did you make any modification to the generated answer? If so when and how.
% What would you like future LiveRAG Challenges to include?
%%%%%%%%%%%%%%%%%%%%%%%%%%%%%%%%%%%%%%%%%%%%

\bibliographystyle{ACM-Reference-Format}
\bibliography{Z-sample-base}

%%% -*-BibTeX-*-
%%% Do NOT edit. File created by BibTeX with style
%%% ACM-Reference-Format-Journals [18-Jan-2012].

\begin{thebibliography}{14}

%%% ====================================================================
%%% NOTE TO THE USER: you can override these defaults by providing
%%% customized versions of any of these macros before the \bibliography
%%% command.  Each of them MUST provide its own final punctuation,
%%% except for \shownote{} and \showURL{}.  The latter two
%%% do not use final punctuation, in order to avoid confusing it with
%%% the Web address.
%%%
%%% To suppress output of a particular field, define its macro to expand
%%% to an empty string, or better, \unskip, like this:
%%%
%%% \newcommand{\showURL}[1]{\unskip}   % LaTeX syntax
%%%
%%% \def \showURL #1{\unskip}           % plain TeX syntax
%%%
%%% ====================================================================

\ifx \showCODEN    \undefined \def \showCODEN     #1{\unskip}     \fi
\ifx \showISBNx    \undefined \def \showISBNx     #1{\unskip}     \fi
\ifx \showISBNxiii \undefined \def \showISBNxiii  #1{\unskip}     \fi
\ifx \showISSN     \undefined \def \showISSN      #1{\unskip}     \fi
\ifx \showLCCN     \undefined \def \showLCCN      #1{\unskip}     \fi
\ifx \shownote     \undefined \def \shownote      #1{#1}          \fi
\ifx \showarticletitle \undefined \def \showarticletitle #1{#1}   \fi
\ifx \showURL      \undefined \def \showURL       {\relax}        \fi
% The following commands are used for tagged output and should be
% invisible to TeX
\providecommand\bibfield[2]{#2}
\providecommand\bibinfo[2]{#2}
\providecommand\natexlab[1]{#1}
\providecommand\showeprint[2][]{arXiv:#2}

\bibitem[Dhole(2024)]%
        {Dhole_PyTerrier_Genrank}
\bibfield{author}{\bibinfo{person}{Kaustubh Dhole}.} \bibinfo{year}{2024}\natexlab{}.
\newblock \bibinfo{booktitle}{\emph{{PyTerrier-GenRank: The PyTerrier Plugin for Reranking with Large Language Models}}}.
\newblock
\urldef\tempurl%
\url{https://github.com/emory-irlab/pyterrier_genrank}
\showURL{%
\tempurl}


\bibitem[Filice et~al\mbox{.}(2025)]%
        {datamorgana_filiceGeneratingDiverseQA2025}
\bibfield{author}{\bibinfo{person}{Simone Filice}, \bibinfo{person}{Guy Horowitz}, \bibinfo{person}{David Carmel}, \bibinfo{person}{Zohar Karnin}, \bibinfo{person}{Liane {Lewin-Eytan}}, {and} \bibinfo{person}{Yoelle Maarek}.} \bibinfo{year}{2025}\natexlab{}.
\newblock \bibinfo{title}{Generating {{Diverse Q}}\&{{A Benchmarks}} for {{RAG Evaluation}} with {{DataMorgana}}}.
\newblock
\href{https://doi.org/10.48550/arXiv.2501.12789}{doi:\nolinkurl{10.48550/arXiv.2501.12789}}
\showeprint[arxiv]{2501.12789}~[cs]


\bibitem[Khattab et~al\mbox{.}(2023)]%
        {khattabDSPyCompilingDeclarative2023}
\bibfield{author}{\bibinfo{person}{Omar Khattab}, \bibinfo{person}{Arnav Singhvi}, \bibinfo{person}{Paridhi Maheshwari}, \bibinfo{person}{Zhiyuan Zhang}, \bibinfo{person}{Keshav Santhanam}, \bibinfo{person}{Sri Vardhamanan}, \bibinfo{person}{Saiful Haq}, \bibinfo{person}{Ashutosh Sharma}, \bibinfo{person}{Thomas~T. Joshi}, \bibinfo{person}{Hanna Moazam}, \bibinfo{person}{Heather Miller}, \bibinfo{person}{Matei Zaharia}, {and} \bibinfo{person}{Christopher Potts}.} \bibinfo{year}{2023}\natexlab{}.
\newblock \bibinfo{title}{{{DSPy}}: {{Compiling Declarative Language Model Calls}} into {{Self-Improving Pipelines}}}.
\newblock
\href{https://doi.org/10.48550/arXiv.2310.03714}{doi:\nolinkurl{10.48550/arXiv.2310.03714}}
\showeprint[arxiv]{2310.03714}~[cs]


\bibitem[Lee(1995)]%
        {lee_combining_1995}
\bibfield{author}{\bibinfo{person}{Joon~Ho Lee}.} \bibinfo{year}{1995}\natexlab{}.
\newblock \showarticletitle{Combining multiple evidence from different properties of weighting schemes}. In \bibinfo{booktitle}{\emph{Proceedings of the 18th annual international {ACM} {SIGIR} conference on {Research} and development in information retrieval}} \emph{(\bibinfo{series}{{SIGIR} '95})}. \bibinfo{publisher}{Association for Computing Machinery}, \bibinfo{address}{New York, NY, USA}, \bibinfo{pages}{180--188}.
\newblock
\showISBNx{978-0-89791-714-8}
\href{https://doi.org/10.1145/215206.215358}{doi:\nolinkurl{10.1145/215206.215358}}


\bibitem[Lin(2004)]%
        {lin2004rouge}
\bibfield{author}{\bibinfo{person}{Chin-Yew Lin}.} \bibinfo{year}{2004}\natexlab{}.
\newblock \showarticletitle{Rouge: A package for automatic evaluation of summaries}. In \bibinfo{booktitle}{\emph{Text summarization branches out}}. \bibinfo{pages}{74--81}.
\newblock


\bibitem[Ma et~al\mbox{.}(2024)]%
        {rankllama}
\bibfield{author}{\bibinfo{person}{Xueguang Ma}, \bibinfo{person}{Liang Wang}, \bibinfo{person}{Nan Yang}, \bibinfo{person}{Furu Wei}, {and} \bibinfo{person}{Jimmy Lin}.} \bibinfo{year}{2024}\natexlab{}.
\newblock \showarticletitle{Fine-Tuning LLaMA for Multi-Stage Text Retrieval}. In \bibinfo{booktitle}{\emph{Proceedings of the 47th International ACM SIGIR Conference on Research and Development in Information Retrieval}} \emph{(\bibinfo{series}{SIGIR 2024})}. \bibinfo{publisher}{ACM}, \bibinfo{pages}{2421–2425}.
\newblock
\href{https://doi.org/10.1145/3626772.3657951}{doi:\nolinkurl{10.1145/3626772.3657951}}


\bibitem[Nogueira et~al\mbox{.}(2019)]%
        {nogueira2019document}
\bibfield{author}{\bibinfo{person}{Rodrigo Nogueira}, \bibinfo{person}{Wei Yang}, \bibinfo{person}{Jimmy Lin}, {and} \bibinfo{person}{Kyunghyun Cho}.} \bibinfo{year}{2019}\natexlab{}.
\newblock \showarticletitle{Document expansion by query prediction}.
\newblock \bibinfo{journal}{\emph{arXiv preprint arXiv:1904.08375}} (\bibinfo{year}{2019}).
\newblock


\bibitem[Opsahl-Ong et~al\mbox{.}(2024)]%
        {opsahl-ong_optimizing_2024}
\bibfield{author}{\bibinfo{person}{Krista Opsahl-Ong}, \bibinfo{person}{Michael~J. Ryan}, \bibinfo{person}{Josh Purtell}, \bibinfo{person}{David Broman}, \bibinfo{person}{Christopher Potts}, \bibinfo{person}{Matei Zaharia}, {and} \bibinfo{person}{Omar Khattab}.} \bibinfo{year}{2024}\natexlab{}.
\newblock \bibinfo{title}{Optimizing {Instructions} and {Demonstrations} for {Multi}-{Stage} {Language} {Model} {Programs}}.
\newblock
\href{https://doi.org/10.48550/arXiv.2406.11695}{doi:\nolinkurl{10.48550/arXiv.2406.11695}}
\newblock
\shownote{arXiv:2406.11695 [cs]}.


\bibitem[Papineni et~al\mbox{.}(2002)]%
        {papineni2002bleu}
\bibfield{author}{\bibinfo{person}{Kishore Papineni}, \bibinfo{person}{Salim Roukos}, \bibinfo{person}{Todd Ward}, {and} \bibinfo{person}{Wei-Jing Zhu}.} \bibinfo{year}{2002}\natexlab{}.
\newblock \showarticletitle{Bleu: a method for automatic evaluation of machine translation}. In \bibinfo{booktitle}{\emph{Proceedings of the 40th annual meeting of the Association for Computational Linguistics}}. \bibinfo{pages}{311--318}.
\newblock


\bibitem[Penedo et~al\mbox{.}(2024)]%
        {penedo2024fineweb}
\bibfield{author}{\bibinfo{person}{Guilherme Penedo}, \bibinfo{person}{Hynek Kydl{\'\i}{\v{c}}ek}, \bibinfo{person}{Anton Lozhkov}, \bibinfo{person}{Margaret Mitchell}, \bibinfo{person}{Colin~A Raffel}, \bibinfo{person}{Leandro Von~Werra}, \bibinfo{person}{Thomas Wolf}, {et~al\mbox{.}}} \bibinfo{year}{2024}\natexlab{}.
\newblock \showarticletitle{The fineweb datasets: Decanting the web for the finest text data at scale}.
\newblock \bibinfo{journal}{\emph{Advances in Neural Information Processing Systems}}  \bibinfo{volume}{37} (\bibinfo{year}{2024}), \bibinfo{pages}{30811--30849}.
\newblock


\bibitem[Pradeep et~al\mbox{.}(2023)]%
        {pradeep_rankzephyr_2023}
\bibfield{author}{\bibinfo{person}{Ronak Pradeep}, \bibinfo{person}{Sahel Sharifymoghaddam}, {and} \bibinfo{person}{Jimmy Lin}.} \bibinfo{year}{2023}\natexlab{}.
\newblock \bibinfo{title}{{RankZephyr}: {Effective} and {Robust} {Zero}-{Shot} {Listwise} {Reranking} is a {Breeze}!}
\newblock
\href{https://doi.org/10.48550/arXiv.2312.02724}{doi:\nolinkurl{10.48550/arXiv.2312.02724}}
\newblock
\shownote{arXiv:2312.02724 [cs]}.


\bibitem[Soylu et~al\mbox{.}(2024)]%
        {soylu_fine-tuning_2024}
\bibfield{author}{\bibinfo{person}{Dilara Soylu}, \bibinfo{person}{Christopher Potts}, {and} \bibinfo{person}{Omar Khattab}.} \bibinfo{year}{2024}\natexlab{}.
\newblock \bibinfo{title}{Fine-{Tuning} and {Prompt} {Optimization}: {Two} {Great} {Steps} that {Work} {Better} {Together}}.
\newblock
\href{https://doi.org/10.48550/arXiv.2407.10930}{doi:\nolinkurl{10.48550/arXiv.2407.10930}}
\newblock
\shownote{arXiv:2407.10930 [cs]}.


\bibitem[Sun et~al\mbox{.}(2024)]%
        {sun_is_2024}
\bibfield{author}{\bibinfo{person}{Weiwei Sun}, \bibinfo{person}{Lingyong Yan}, \bibinfo{person}{Xinyu Ma}, \bibinfo{person}{Shuaiqiang Wang}, \bibinfo{person}{Pengjie Ren}, \bibinfo{person}{Zhumin Chen}, \bibinfo{person}{Dawei Yin}, {and} \bibinfo{person}{Zhaochun Ren}.} \bibinfo{year}{2024}\natexlab{}.
\newblock \bibinfo{title}{Is {ChatGPT} {Good} at {Search}? {Investigating} {Large} {Language} {Models} as {Re}-{Ranking} {Agents}}.
\newblock
\href{https://doi.org/10.48550/arXiv.2304.09542}{doi:\nolinkurl{10.48550/arXiv.2304.09542}}
\newblock
\shownote{arXiv:2304.09542 [cs]}.


\bibitem[Team(2024)]%
        {Falcon3}
\bibfield{author}{\bibinfo{person}{Falcon-LLM Team}.} \bibinfo{year}{2024}\natexlab{}.
\newblock \bibinfo{title}{The Falcon 3 Family of Open Models}.
\newblock
\urldef\tempurl%
\url{https://huggingface.co/blog/falcon3}
\showURL{%
\tempurl}


\end{thebibliography}

%% If your work has an appendix, this is the place to put it.
\appendix

\end{document}